\documentclass[lettersize,journal,fleqn]{IEEEtran}
\usepackage{amsmath,amsfonts}
\usepackage{algorithmic}
\usepackage{algorithm}
\usepackage[algo2e]{algorithm2e} 
\usepackage{array}
\usepackage[caption=false,font=normalsize,labelfont=sf,textfont=sf]{subfig}
\usepackage{textcomp}
\usepackage{stfloats}
\usepackage{url}
\usepackage{verbatim}
\usepackage{graphicx}
\usepackage{cite}
\usepackage{eucal}
\usepackage{mathrsfs}
\usepackage{courier}
\usepackage{xcolor,soul}

\usepackage{multirow}
\usepackage[export]{adjustbox}

\begin{document}

\title{Crown-CAM: Interpretable Visual Explanations for \hspace{0.5cm} Tree Crown Detection in Aerial Images}

\author{Seyed Mojtaba Marvasti-Zadeh,~\IEEEmembership{Member,~IEEE}, Devin Goodsman, Nilanjan Ray,~\IEEEmembership{Member,~IEEE}, Nadir Erbilgin
\thanks{This work was supported by the fRI Research - Mountain Pine Beetle Ecology Program.\\
S.M. Marvasti-Zadeh and N. Erbilgin are with the Department of Renewable Resources, University of Alberta, Canada (\{mojtaba.marvasti, erbilgin\}@ualberta.ca), 
D. Goodsman is with the Canadian Forest Service, Natural Resources Canada, Canada (devin.goodsman@nrcan-rncan.gc.ca), and N. Ray is with the Department of Computing Science, University of Alberta, Canada (nray1@ualberta.ca).
}
\thanks{The manuscript has been accepted in IEEE Geoscience and Remote Sensing Letters (GRSL).}
}
\markboth{Journal of \LaTeX\ Class Files,~Vol.~, No.~, April~2023}%
{Shell \MakeLowercase{\textit{et al.}}: A Sample Article Using IEEEtran.cls for IEEE Journals}

\maketitle
\begin{abstract}
Visual explanation of ``black-box'' models allows researchers in \textit{explainable artificial intelligence} (XAI) to interpret the model's decisions in a human-understandable manner. In this paper, we propose \textit{interpretable class activation mapping for tree crown detection} (Crown-CAM) that overcomes inaccurate localization \& computational complexity of previous methods while generating reliable visual explanations for the challenging and dynamic problem of tree crown detection in aerial images. It consists of an unsupervised selection of activation maps, computation of local score maps, and non-contextual background suppression to efficiently provide fine-grain localization of tree crowns in scenarios with dense forest trees or scenes without tree crowns. Additionally, two \textit{Intersection over Union} (IoU)-based metrics are introduced to effectively quantify both the accuracy and inaccuracy of generated explanations with respect to regions with or even without tree crowns in the image. 
Empirical evaluations demonstrate that the proposed Crown-CAM outperforms the Score-CAM, Augmented Score-CAM, and Eigen-CAM methods by an average IoU margin of 8.7, 5.3, and 21.7 (and 3.3, 9.8, and 16.5) respectively in improving the accuracy (and decreasing inaccuracy) of visual explanations on the challenging NEON tree crown dataset.
\end{abstract}
\section{Introduction}
\IEEEPARstart{E}{xplainable} \textit{artificial intelligence} (XAI) consists of methods that allow humans to describe, interpret, and understand the predictions of ``black-box'' models (e.g., deep neural networks) to ensure end-users can trust the model to provide accurate decisions. In general, XAI methods can be classified based on their interpretability (e.g., perceptive or mathematical structures), methodology (e.g., gradient-based or gradient-free), and model usage (e.g., model-specific or post-hoc) \cite{Survey_XAI_Opportunity,Survey_XAI_Medical}. In particular, post-hoc perceptive gradient-free methods (e.g., proposed Crown-CAM) are of interest due to their advantages, including being well-designed for \textit{convolutional neural networks} (CNNs), without affecting the model's performance or training process, applicable to models with non-differentiable outputs, and compatible with human perception. Meanwhile, a widely-adopted approach is \textit{class activation mapping} (CAM) to generate a saliency map (or heat-maps) highlighting which features of a given input significantly affect the final decision of the model. \\
\begin{figure}[!t]
\centering
\subfloat[]{\includegraphics[width=0.22\linewidth]{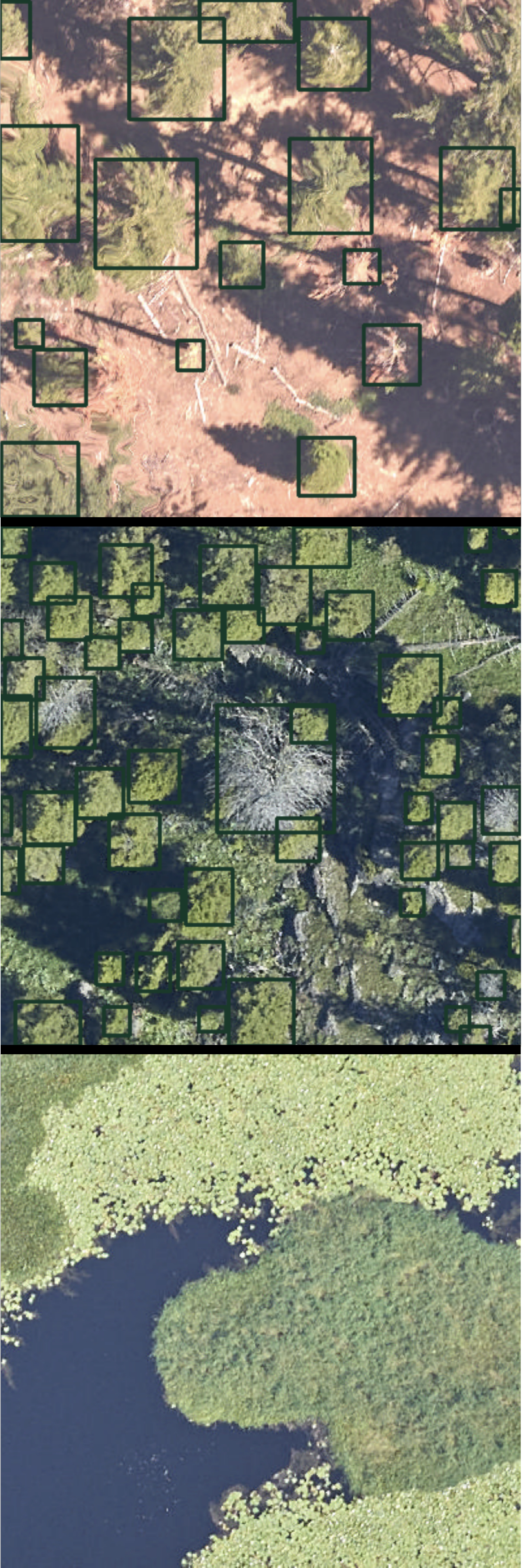}%
\label{fig_bboxes}}
\hfil
\subfloat[]{\includegraphics[width=0.22\linewidth]{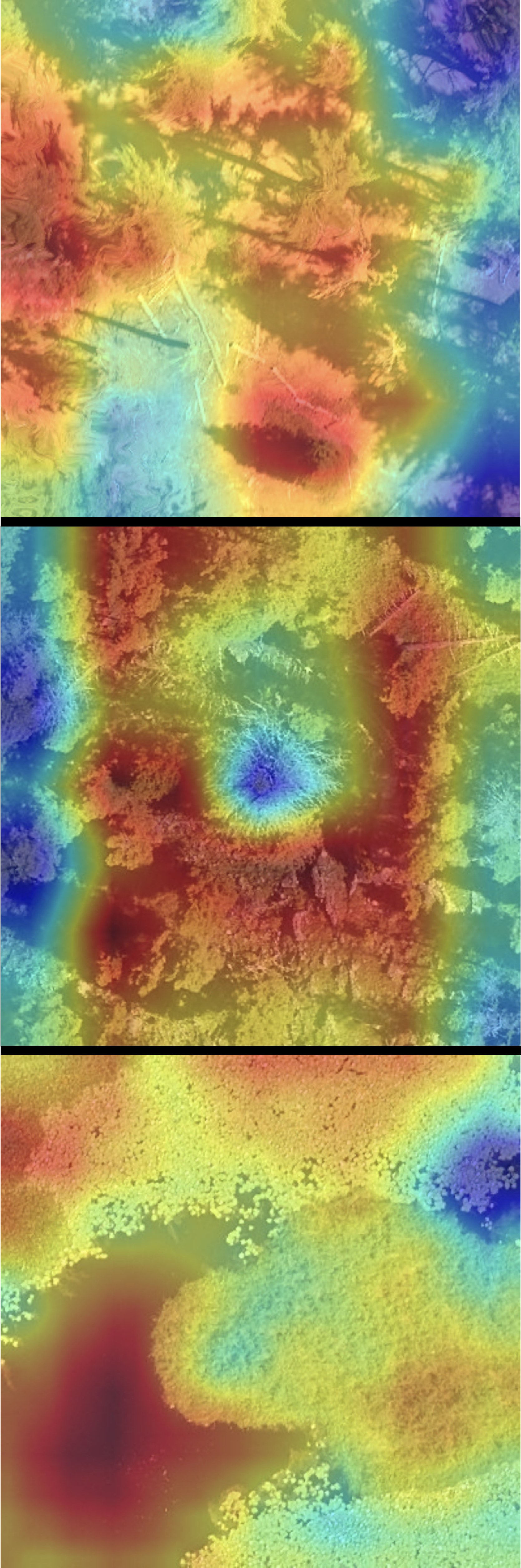}%
\label{fig_scorecam}}
\hfil
\subfloat[]{\includegraphics[width=0.22\linewidth]{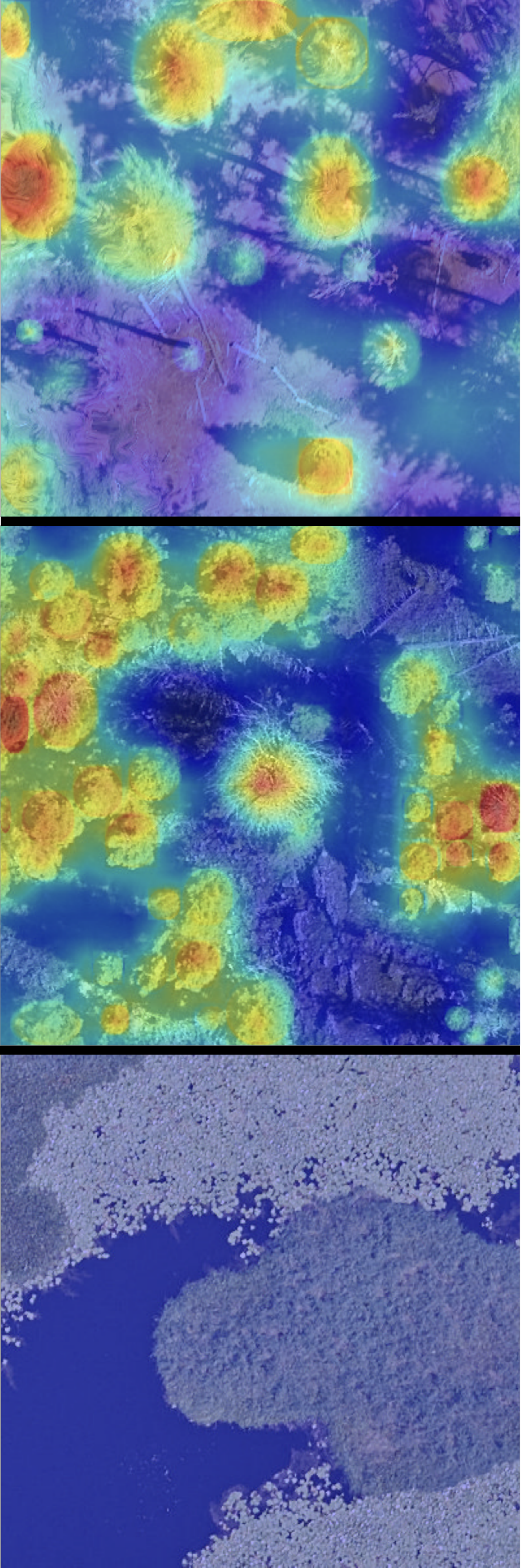}%
\label{fig_crowncam}}
\hspace{.7cm}
\subfloat[]{\includegraphics[width=0.219\linewidth]{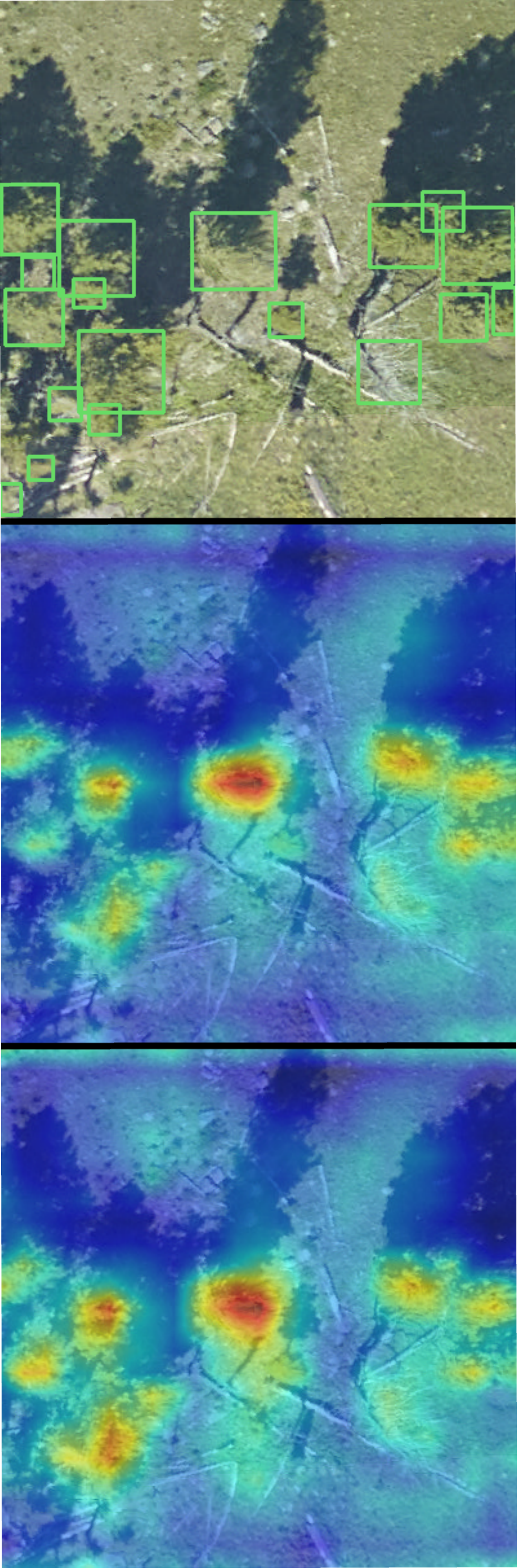}%
\label{fig_channel_reduction}}
\caption{Comparison of (a-c): CAMs generated for aerial image-based tree crown detection scenarios and (d): impact of removing redundant activation maps. Here, (a) illustrates the detected crowns by DeepForest model \cite{DeepForest}, (b) shows the CAMs by Score-CAM \cite{ScoreCAM}, (c) presents the CAMs generated by proposed Crown-CAM, and (d) shows the detected tree crowns (top row) and CAMs generated from all (middle row) and a half (bottom row) of the activation maps. Our Crown-CAM efficiently addresses the coarse localization problem of Score-CAM (e.g., top two rows of (c) \& bottom row of (d)) while suppressing background channels of CAMs (e.g., the bottom row of (c)).}
\label{fig:motivation}
\end{figure}
While the demand for interpretability of deployed AI models via the XAI framework is becoming an unavoidable task in computer vision, explainability remains an emerging concept in solving some complex problems. Tree crown detection is one of these challenging problems in forest remote sensing. It is the critical first step that enables ecologists, foresters, biologists, and land managers to perform tasks such as monitoring vegetation changes, detecting insect infestations, estimating tree density, and identifying tree species. However, the usage of XAI methods in detecting tree crowns has not been investigated despite their wide application for generic object classification. Further, in contrast to XAI applications in classification that mainly involve a few large objects, tree crown detection from aerial images involves challenging scenarios including dense forests with highly overlapping tree crowns, small tree crowns, and potential distractors. Despite the overlap between the generated CAM and the estimated bounding boxes, the visual explanations can provide beyond contextual information (i.e., information that is not directly related to the task, e.g., to refine noisy predictions further) or support in-depth analysis of the overall process (i.e., steps involved in analyzing data, such as preprocessing, feature extraction, model training, prediction, and post-processing). Based on these motivations, this work aims to provide interpretable visual explanations and relevant quantitative evaluation metrics for tree crown detection in aerial images.

This work is motivated by the simple yet effective Score-CAM method \cite{ScoreCAM}, which generates CAMs with promising accuracy and robustness that is independent of gradient computation. It uses convolutional activation maps as masks on the input image and then passes them to the network to compute one score for each activation map. At last, CAM for a class is generated by the weighted sum of activation maps using normalized scores. The Score-CAM has the main disadvantage of coarse localization of generated CAMs \cite{SSCAM,ISCAM} which prevents it from performing well in detecting tree crowns from aerial views (see Fig.~\ref{fig_scorecam}). Several extensions of Score-CAM have been proposed to obtain more accurate and visually sharper CAMs; however, these methods significantly increase computational complexity by combining Score-CAM's pipeline with additional operations or iterations. For instance, the SS-CAM \cite{SSCAM}, IS-CAM \cite{ISCAM}, and Augmented Score-CAM \cite{AugmentedScoreCAM}, respectively, utilize smooth operation, integration operation, and augmented activation maps to reduce inaccurate localization or capture more parts of an object. Meanwhile, the effectiveness of these methods has been evaluated based on qualitative metrics that are both designed for the classification task and focus solely on the objects within a scene. However, these metrics lack the ability to assess the localization accuracy of CAMs in detecting objects and fail to recognize the overlap (or the inaccuracy) between the generated CAMs and the background pixels (e.g., when there is no object in a scene). 
\begin{figure*}[!t]
\subfloat[]{\includegraphics[width=0.781\linewidth,valign=c]{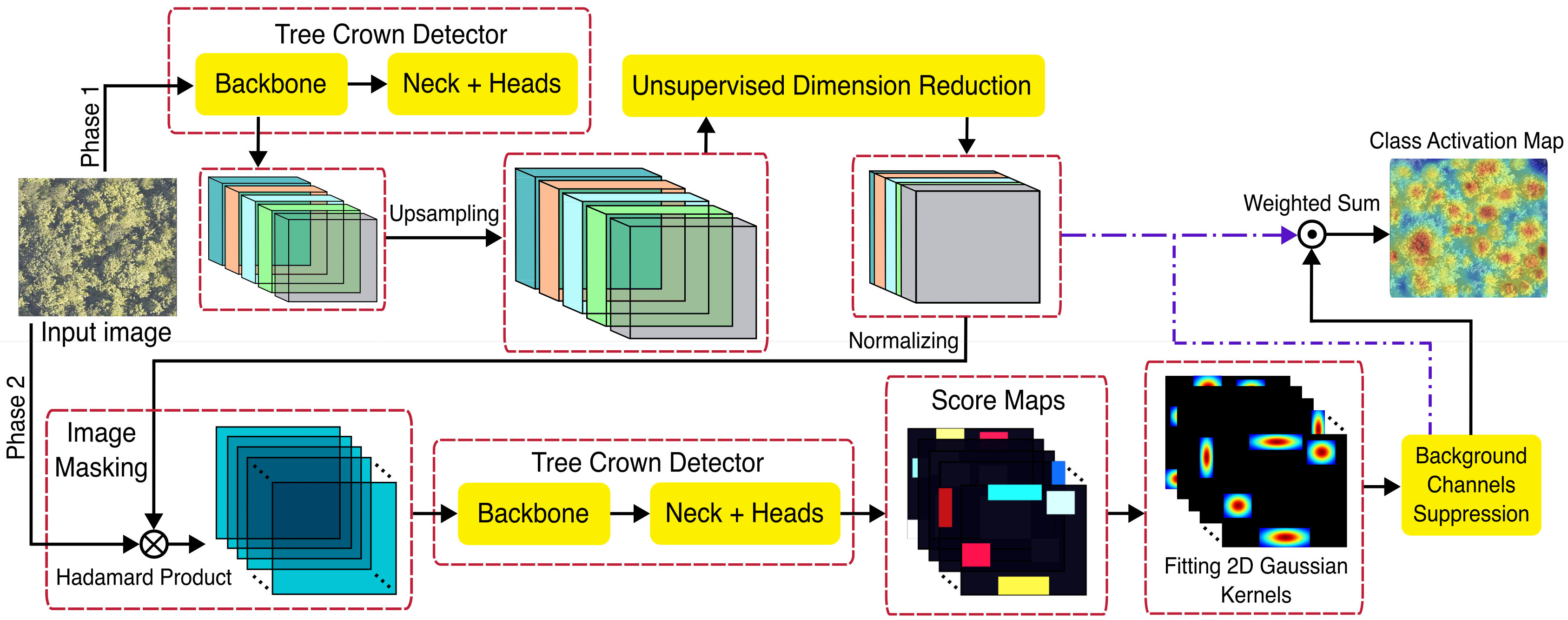}%
\label{fig:pipeline}}
\hspace{.23cm}
\centering
\subfloat[]{\includegraphics[width=0.195\linewidth,valign=c]{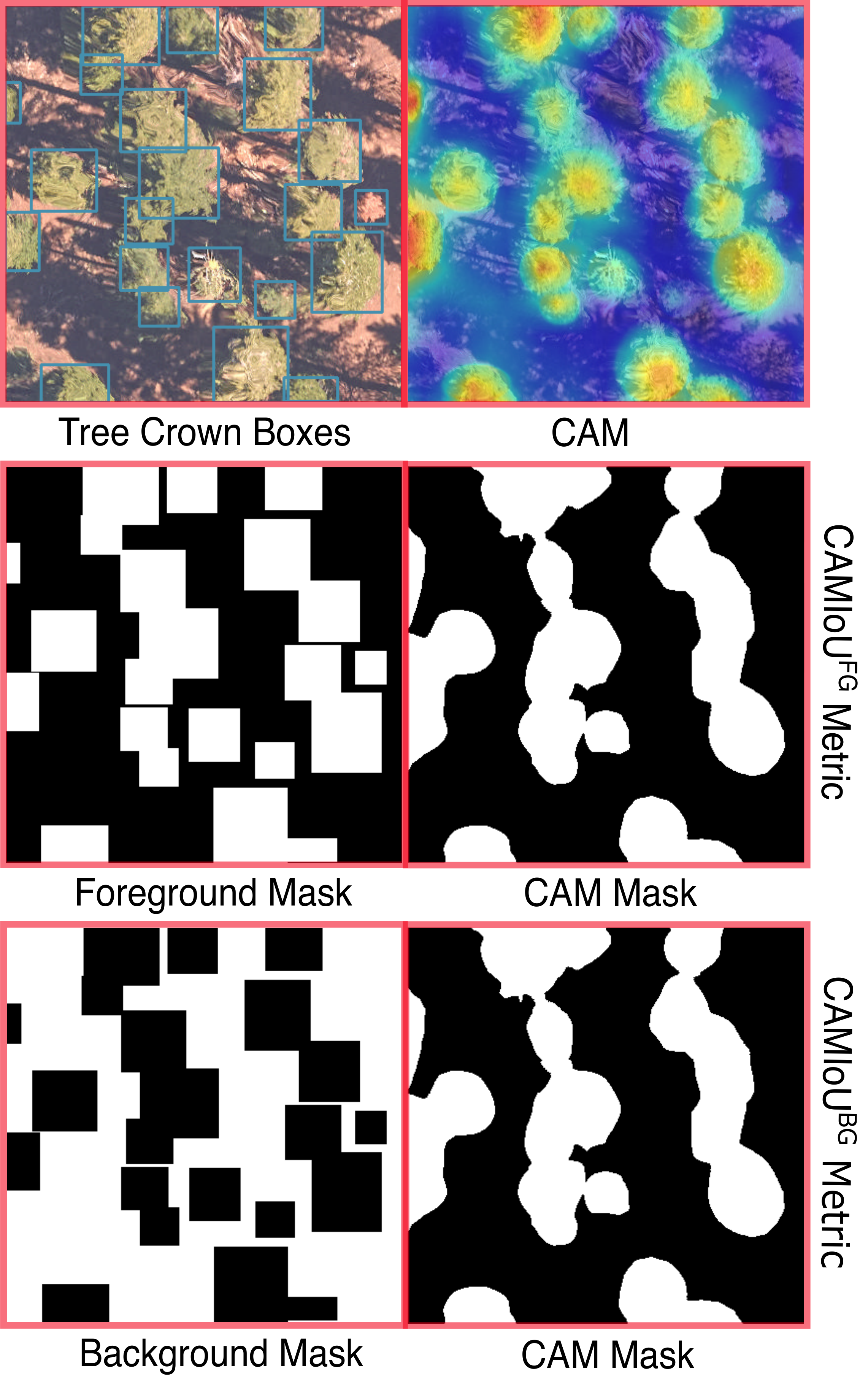}%
\label{fig:metric_masks}}
\caption{An overview of (a) proposed Crown-CAM and (b) generated masks for introduced IoU-based localization metrics. (a) In the first phase, an input image is fed into a well-trained detector to extract the activation of backbone layers, tree crown bounding boxes, and their confidence scores. Next, the proposed unsupervised dimension reduction is used to retain information-rich activation maps and discard redundant ones. In the second phase, the input image is masked by the normalized selected activation maps, and then the masked images are passed into the detector to calculate local score maps. Further, these maps are fitted with 2D Gaussian kernels to avoid boundary effects. Following that, the score map channels corresponding to the background are suppressed and excluded from the selected activation maps (violet dash arrow). Finally, the weighted sum of Gaussian fitted score maps and selected activation maps generate the tree crown CAM. (b) Based on the generated CAM and the bounding boxes of tree crowns, three masks are generated to calculate the localization accuracy (i.e., overlap with tree crowns) and inaccuracy (i.e., overlap with background regions) of visual explanations.}
\end{figure*}
In contrast, we propose Crown-CAM, which provides accurate \& interpretable CAMs for tree crown detection models without considerably increasing the computational burden (see Fig.~\ref{fig_crowncam} \& Fig.~{\ref{fig_channel_reduction}}). It is particularly helpful in real-world scenarios involving various environmental factors where accurate and reliable analysis is essential for environmental monitoring, forest management, and land-use planning, as it can allow remote sensing experts to better interpret \& incorporate the predictions made by trained models. It is also the preferred choice for practical remote sensing applications requiring the ability to process large-scale datasets efficiently. Moreover, inspired by \textit{weakly-supervised localization} (WSL) of CAMs (e.g., \cite{GradCAM}), two metrics are introduced that efficiently quantify the accuracy \& inaccuracy of visual explanations with respect to the tree crowns and background regions. In the WSL, the IoU metric is calculated between the ground-truth bounding box and a bounding box drawn around the single largest segment of binarized CAM. However, we overcome its limitations by extending this metric to the tree crown detection task involving multiple targets with varying confidence scores and efficiently calculating the IoU between a collection of tree crown bounding boxes and a segmented CAM. We also propose an additional metric that calculates the inaccuracy of CAMs when they drift into background regions or are generated in the absence of tree crowns. \\
\indent In summary, the main contributions are as follows.
\begin{itemize}
    \item Our proposed Crown-CAM makes it possible to generate interpretable CAMs for detecting tree crowns in aerial images.
    \item We propose unsupervised dimension reduction of activation maps to provide computational efficiency by excluding redundant activation maps and emphasizing informative ones (see Fig.~\ref{fig_channel_reduction}).
    \item The proposed Crown-CAM computes local score maps rather than one global score per activation map used in prior efforts. Hence, area-based scores merely enable target-related areas of activation maps to provide more accurate CAMs for challenging aerial detection scenarios. In addition, 2D Gaussian kernels are fitted to tree crown areas (i.e., weighting pixels of tree crown areas based on their distance from the center of same-sized Gaussian kernels) to reduce the boundary effects of local scores.
    \item We propose a background channel suppression strategy based on the calculated score maps to avoid misleading activation maps that do not locate the tree crowns.
    \item Two useful IoU-based metrics are introduced that quantify the accuracy (and inaccuracy) of the generated CAM in comparison with dense tree crowns (or background areas). 
\end{itemize}
The rest of the paper is organized as follows. The proposed Crown-CAM and localization metrics are described in detail in Section~\ref{sec:crowncam} and Section~\ref{sec:metrics}. In Section~\ref{sec:experiments}, empirical evaluations and comparisons are presented. Finally, the conclusion is summarized in Section~\ref{sec:conclusion}.
\section{Proposed Method: Crown-CAM}\label{sec:crowncam}
In this section, we introduce Crown-CAM, which enables explanation visualization of tree crown detectors in challenging aerial scenarios. An overview of the proposed method is presented in Fig.~\ref{fig:pipeline}. 
The Crown-CAM comprises two main phases: i) extracting activation maps from all backbone layers of a tree crown detector (and tree crown bounding boxes \& their confidence scores), and unsupervised dimension reduction, and ii) image masking, local score map calculation, fitting 2D Gaussian kernels to the score maps, and background channels suppression. A pseudo-code for our Crown-CAM is shown in Algorithm~\ref{algorithm}. Each step is described in detail in the following subsections.
\subsection{Phase $\mathit{1}$: Unsupervised Selection of Activation Maps}
In this phase, we seek to minimize the computational complexity of high-dimensional activation maps required to generate CAMs for the detection task (in contrast to previous cumbersome classification methods requiring extra operations and iterations \cite{SSCAM,ISCAM,AugmentedScoreCAM}). For a given input image $X \in \mathbb{R}^{3\times w \times h}$ ($w$ and $h$ refer to width and height), we can use a well-trained tree crown detector $f$ to extract activation maps $\mathbb{A}$ from backbone layers $l \in [1,\dots,\rm{L}]$ (each having $C$ channels) in addition to detected bounding boxes $B^{X}$ and corresponding confidence scores $Y^{X}$. When using XAI for classification tasks, a widely-adopted technique is to extract activation maps only from the last backbone layer, thus reducing the computational burden. However, the spatial resolution of these activation maps is too coarse to generate accurate CAMs for challenging tree crown detection scenarios, requiring pixel-accurate localization. Hence, the Crown-CAM involves the activation maps of all backbone layers to generate interpretable CAMs that incorporate fine-grained tree crown information. 

Next, the extracted activation maps are upsampled (i.e., $\mathbb{U} \in \mathbb{R}^{(L\times C)\times w \times h}$) by applying bi-linear interpolation to match the input image size. The baseline Score-CAM \cite{ScoreCAM} (and its extensions \cite{SSCAM,ISCAM,AugmentedScoreCAM}) perturbs an input image with these activation maps and calculates their importance. However, not all of the extracted activation maps contribute significantly to the generation of CAM due to their high correlations and redundancies (see Fig.~\ref{fig_channel_reduction}). Meanwhile, the most informative activation maps should be retained to ensure that the visual explanations generated are as accurate as possible. Accordingly, a simple but effective unsupervised dimension reduction is proposed using KL-divergence loss between each activation map and all other maps, defined as
\begin{flalign}\label{eq:kl}
\small
\rm{{D}_{KL}^\mathit{c} = KL}(\mathbb{U}^\mathit{c} \parallel \mathbb{U}^{\mathbb{C}\setminus{\{\mathit{c}\}}}),\hspace{1.1cm} \rm{\mathit{c} \in {[1,\ldots, {L\times C}]}}
\end{flalign}
where the lower values correspond to redundant activation maps. In other words, it aims to select activation maps $\bar{\mathbb{U}} \in \mathbb{R}^{\bar{C}\times w \times h}$ that minimize information loss and ensure computation efficiency. Hence, the number of activation maps was experimentally reduced by a factor of two (i.e., $\bar{C}=1/2(L\times C)$). 
This factor was achieved by qualitative comparison (provided a more intuitive understanding of the model's predictions) of CAMs generated with the pre-trained DeepForest model \mbox{\cite{DeepForest}} (see Fig.~\mbox{\ref{fig_channel_reduction}}), which was then applied to different detectors.
Afterwards, the selected activation maps with maximal KL-divergence loss will be normalized to serve as masks $\mathbb{N}$ for the next phase.
\subsection{Phase $\mathit{2}$: Local Score Maps for Activation Maps}
The first step in this phase is to mask the input image by applying the normalized activation maps as
\begin{flalign}\label{eq:masking}
\small
\rm{\bar{X}^\mathit{c} = \mathbb{N}^\mathit{c} \circ \rm{X} },\hspace{3.6cm} \rm{\mathit{c} = {[1,\ldots, \bar{C}]}}
\end{flalign}
where $\circ$ indicates the Hadamard product. The masked images $\bar{X}$ are then passed into the tree crown detector $f$ to calculate the local score maps $S$. While the baseline Score-CAM (and its modified versions \cite{SSCAM,ISCAM,AugmentedScoreCAM}) calculates a global score for each activation map, the proposed Crown-CAM provides a score map per activation map that is locally weighted based on the predicted bounding boxes and their confidence scores as 
\begin{flalign}\label{eq:scores}
\small
\rm{S_\mathit{k}^\mathit{c} = }~~   IoU(\rm{B_\mathit{k}^X, B_\mathit{k}^{\bar{X}^\mathit{c}}}) + \mid Y_\mathit{k}^{{X}} - Y_\mathit{k}^{\bar{X}^\mathit{c}}\mid ,
\end{flalign}
in which $B^{\bar{X}^c}$, $Y^{\bar{X}^c}$, and $k$ are the predicted bounding boxes, their confidence scores, and the index of predicted boxes, respectively. Accordingly, important areas of each score map are weighted using the IoU between the predicted bounding boxes derived from the input image and the activation-based image masks as well as the absolute difference between their confidence scores. 
However, applying weights to these areas may cause boundary effects, meaning that the weights assigned to pixels near the boundaries of these areas may appear inconsistent with adjacent pixels.
To alleviate these unwanted artifacts and smooth the boundaries of tree crown areas in the local score maps, 2D Gaussian kernels with $\bar{w}_{k}\times\bar{h}_{k}$ dimensions are fitted to the center of predicted tree crowns with variance $\sigma^2$ as
\begin{flalign}\label{eq:gauss}
\small
\rm{ \mathbb{Z}_\mathit{k}^\mathit{c}(B_\mathit{k}^{\bar{X}^\mathit{c}}) = S_\mathit{k}^\mathit{c} \cdot \mathcal{N}(B | B_\mathit{k}^{\bar{X}^\mathit{c}}, \sigma^2) }
\end{flalign}
\noindent where $\mathbb{Z}\in \mathbb{R}^{\bar{C}\times w \times h}$ and $\mathcal{N}$ are zero maps and 2D Gaussian function, respectively. 
\begin{algorithm}[H]
\caption{Pseudo-code of Crown-CAM}\label{alg:alg1}
{\textsc{\textbf{\scriptsize Inputs:}}} \scriptsize Input image $\rm{X}$, Detection model $\mathit{f(\cdot)} $, Backbone layers $\mathit{l} \in [1,\ldots,\rm{L}]$, Desired number of channels $ \rm{\bar{C}} $ \\
{\textsc{\textbf{Output:}}} Class activation map $\rm{M_{CrownCAM}}$ \\
{\textsc{\textbf{Notations:}}} Bounding boxes $\rm{B}$, Confidence scores $\rm{Y}$, Activation of layers $\rm{\mathbb{A}}$, KL loss $\rm{D_{KL}}$, IoU metric $\rm{IoU(\cdot,\cdot)}$, Local score maps $\rm{S}$, Gaussian-fitted score maps $\rm{\Delta}$, Activation and local score maps after channel suppression $\rm{\hat{\mathbb{U}}}$ and $\rm{\hat{\mathbb{Z}}}$ \\
$ \rm{ B^{X}, Y^{X}, \mathbb{A}_{\ell} }  \gets \mathit{f}(X) $  \hspace{3.2cm} \textcolor{teal}{$\triangleright$ \texttt{\scriptsize Apply model to image}}  \\
\For{$\ell$ $\rm{in}$ $\rm{[1,\ldots, L]}$}{
    \For{$c$ $\rm{in}$ $\rm{[1,\ldots, {C}]}$}{ 
         $\mathbb{U}_\ell^\mathit{c} \gets$ Up-sample $\rm{{\mathbb{A}}_\ell^\mathit{c}}$ 
    }
}
\For{$c$ $\rm{in}$ $\rm{[1,\ldots, {L\times C}]}$}{ 
    $ \rm{{D}_{KL}^\mathit{c} = KL}(\mathbb{U}^\mathit{c} \parallel \mathbb{U}^{\mathbb{C}\setminus{\{\mathit{c}\}}}) $\hspace{2.54cm} \textcolor{teal}{$\triangleright$ \texttt{\scriptsize Unsupervised loss}}
}
$ \bar{\mathbb{U}} \gets $ Select $\rm{\bar{{C}}}$ indices of $\mathbb{U}$ with maximal $\rm{{D}_{KL}}$ \\
\For{$c$ $\rm{in}$ $\rm{[1,\ldots, \bar{{C}}]}$}{ 
    $ \mathbb{N}^c \gets $ Normalize $\bar{\mathbb{U}}^c$ \\
    $ \rm{\bar{X}^\mathit{c}} \gets \mathbb{N}^\mathit{c} \circ \rm{X} $ \hspace{4cm} \textcolor{teal}{$\triangleright$ \texttt{\scriptsize Hadamard product}} \\ 
    $ \rm{ B^{\bar{X}^\mathit{c}}, Y^{\bar{X}^\mathit{c}}}  \gets \mathit{f}(\bar{X}^\mathit{c}) $ \hspace{1.5cm} \textcolor{teal}{$\triangleright$ \texttt{\scriptsize Apply model to masked images}} \\ 
    \For{$k$ $\rm{in}$ $\rm{[1,\ldots, length(Y^{\bar{X}^\mathit{c}})]}$}{ 
        $ \rm{S_\mathit{k}^\mathit{c} =} $ IoU$(\rm{B_\mathit{k}^X, B_\mathit{k}^{\bar{X}^\mathit{c}}}) + \mid Y_\mathit{k}^{{X}} - Y_\mathit{k}^{\bar{X}^\mathit{c}} \mid $ \hspace{1.75cm} \textcolor{teal}{$\triangleright$ \texttt{\scriptsize Score maps}}\\
        $ \rm{ \mathbb{Z}_\mathit{k}^\mathit{c}(B_\mathit{k}^{\bar{X}^\mathit{c}}) = S_\mathit{k}^\mathit{c} \cdot \mathcal{N}(B | B_\mathit{k}^{\bar{X}^\mathit{c}}, \sigma^2) } $ \hspace{1.9cm} \textcolor{teal}{$\triangleright$ \texttt{\scriptsize Fit Gaussians}}
    }
    Suppress $ \rm{ \mathbb{Z}^\mathit{c}, \bar{\mathbb{U}}^\mathit{c}} $ \textit{iff} $ \sum_{\mathit{\kappa}}{\mathbb{Z}_\mathit{\kappa}^\mathit{c}}=0 $ 
}
$ \rm{ {\Delta}^\mathit{c} \gets \frac{\exp(\hat{\mathbb{Z}}^\mathit{c})}{\sum_{\mathit{c=1}}^{\hat{C}}{\exp(\mathbb{\hat{Z}}^\mathit{c})}}}$\hspace{0.05cm}, \hspace{0.05cm} $\sum_{\mathit{c}}{\Delta^\mathit{c}}=1 $ \textcolor{teal}{\hspace{1.1cm} $\triangleright$ \texttt{\scriptsize Scores to probabilities}} \\
$ \rm{ M_{CrownCAM} \gets ReLU(\sum_{\mathit{c}}{{\Delta}_\mathit{c} \cdot \hat{\mathbb{U}}_\mathit{c}}) } $
\label{algorithm}
\end{algorithm}

The baseline Score-CAM and its modified versions apply softmax (i.e., ${\exp(\rho)}/{\sum_{\mathit{\delta}}{\exp(\rho)}}$) to the global score vector of activation maps to ensure that the sum of values is one (i.e., probability interpretation). However, background channels may adversely affect the generation of CAMs in these methods. Here is a simple example that clarifies the issue with a score vector of length 10. If no object of interest is present in the input image, then all elements in Score-CAM's score vector will be zero. However, applying softmax to the zero vector will result in 0.1 per its elements. In other words, the CAM will be generated based on the weighted sum of all activation maps with non-zero factors of 0.1, thus allowing non-target-related areas to be activated (see the bottom row of Fig.~\ref{fig_scorecam}). To address this problem, the proposed Crown-CAM suppresses background channels if all elements of local score maps are zero. Thus, it minimizes the impact of false background regions on the final generated CAM (see the bottom row of Fig.~\ref{fig_crowncam}). Finally, the softmax is applied on the remaining local score maps (i.e., $\hat{\mathbb{Z}}$), and the final CAM is obtained by 
\begin{flalign}\label{eq:cam}
\small
\rm{ M_{CrownCAM}=ReLU(\sum\nolimits_{\mathit{c}}{{\Delta}_\mathit{c} \cdot \hat{\mathbb{U}}_\mathit{c}}) }
\end{flalign}
where $\Delta$ represents softmax-applied local score maps and ReLU is used to remove activations with negative effects.
\begin{table*}[b!]
\scriptsize
\caption{Quantitative comparison of gradient-free visual explanation methods in terms of proposed metrics for tree crown detection.} 
\centering 
\resizebox{\textwidth}{!}{
\begin{tabular}{c | c c | c c | c c | c c} 
\hline \hline
\multirow{2}{*}{Tree Crown Detector} & \multicolumn{2}{c|}{Proposed Crown-CAM} & \multicolumn{2}{c|}{Score-CAM \cite{ScoreCAM}} & \multicolumn{2}{c|}{Augmented Score-CAM \cite{AugmentedScoreCAM}}  & \multicolumn{2}{c}{Eigen-CAM \cite{EigenCAM}}  \\
& \texttt{CAMIoU$^{\texttt{FG}}$} ($\uparrow$) & \texttt{CAMIoU$^{\texttt{BG}}$} ($\downarrow$) & \texttt{CAMIoU$^{\texttt{FG}}$} ($\uparrow$) & \texttt{CAMIoU$^{\texttt{BG}}$} ($\downarrow$) & \texttt{CAMIoU$^{\texttt{FG}}$} ($\uparrow$) & \texttt{CAMIoU$^{\texttt{BG}}$} ($\downarrow$) & \texttt{CAMIoU$^{\texttt{FG}}$} ($\uparrow$) & \texttt{CAMIoU$^{\texttt{BG}}$} ($\downarrow$) \\\hline 
DeepForest \cite{DeepForest}  & \textbf{29.64} & \textbf{19.81} & 24.96 & 23.37 & 25.64 & 24.89 & 14.91 & 32.28 \\\hline
Faster-RCNN  & \textbf{26.13} & \textbf{25.10} & 21.72 & 34.98 & 23.41 & 31.26 & 5.26 & 34.60  \\\hline
Cascade-RCNN  & \textbf{42.47} & 6.18 & 25.44 & 2.72 & 33.06 & 4.79 & 12.80 & \textbf{0.77}  \\\hline
\hline
\end{tabular}}
\label{table:comparison}
\end{table*}
\section{Proposed Metrics for Evaluation}\label{sec:metrics}
For the classification task, CAMs are evaluated by either the percentage drop (or increase) in the model's score when CAM is provided as input (i.e., Drop (or Increase) in Confidence metrics) or the number of times the Drop-in-Confidence metric is less for one method than in another (i.e., Win metric). However, these metrics do not exploit the contextual information in a scene and therefore fail to deal with the challenges involved in tree crown detection. Moreover, for the WSL setting, the object localization is calculated based on a labeled bounding box and a rectangle drawn around the largest segment of interest. However, it has not been extended to dense object detection, cannot evaluate the inaccuracy of generated CAMs without objects of interest, and cannot quantify dense boxes simultaneously. To address these limitations, we introduce two IoU-based metrics, \texttt{CAMIoU$^{\texttt{FG}}$} and \texttt{CAMIoU$^{\texttt{BG}}$}, which indicate the accuracy of CAMs in detecting tree crowns and avoiding background regions. These metrics are calculated by generating three masks based on the CAM and bounding boxes of the tree crowns as
\begin{flalign}\label{eq:mask_cam}
\small
\rm{{Mask}^{\rm{CAM}} =} 
\begin{cases}
    1,& \mathit{if} \hspace{0.15cm} \rm{CAM(i,\hspace{0.05cm} j)}\geq \rm{Threshold}\\
    0,              & \mathit{otherwise}
\end{cases}
\end{flalign}
\begin{flalign}\label{eq:mask_fg_bg}
\small
\rm{{Mask}^{\rm{FG \hspace{0.05cm} (or\hspace{0.1cm}BG)}} =} 
\begin{cases}
    1 \hspace{0.15cm}\rm{(or \hspace{0.15cm}0)},& \mathit{if } \hspace{0.15cm} \rm{\zeta(i, \hspace{0.05cm} j)}\in \rm{\mathbb{B}^{\rm{X}}}\\
    0 \hspace{0.15cm}\rm{(or \hspace{0.15cm}1)},              & \mathit{otherwise}
\end{cases}
\end{flalign}
where $\zeta\in \mathbb{R}^{w\times h}$ refers to a zeros (or ones) map with spatial locations of $(i,\hspace{0.05cm}j)$. Also, ${\mathbb{B}^{{X}}}$ indicates the ground-truth tree crown bounding boxes (i.e., tree crown boxes in the testing set). Then, the localization metrics can be defined as follows.
\begin{flalign}\label{eq:camiou_fg}
\small
\texttt{CAMIoU$^{\texttt{FG}}$ =} 
\frac{\rm{ {Mask}^{\rm{CAM}} }\hspace{0.05cm}\texttt{.}\hspace{0.05cm}  \rm{ {Mask}^{\rm{FG}}} }{\rm{ {Mask}^{\rm{CAM}}} \hspace{0.05cm}\texttt{+}\hspace{0.05cm} \rm{ {Mask}^{\rm{FG}}}}
\end{flalign}
\begin{flalign}\label{eq:camiou_bg}
\small
\texttt{CAMIoU$^{\texttt{BG}}$ =} 
\frac{\rm{ {Mask}^{\rm{CAM}}}\hspace{0.05cm}\texttt{.} \hspace{0.05cm}  \rm{ {Mask}^{\rm{BG}}} }{\rm{ {Mask}^{\rm{CAM}}} \hspace{0.05cm}\texttt{+}\hspace{0.05cm} \rm{ {Mask}^{\rm{BG}}}}
\end{flalign}
Here, the numerators represent pixel-based overlap areas between salient regions of CAM and tree crowns or backgrounds, while the denominators indicate their union areas.
Therefore, the most interpretable approaches must focus on tree crown regions (i.e., higher \texttt{CAMIoU$^{\texttt{FG}}$}) while avoiding background distractors (i.e., lower \texttt{CAMIoU$^{\texttt{BG}}$}).
\section{Empirical Evaluations}\label{sec:experiments}
In this section, the effectiveness of the proposed Crown-CAM is evaluated using the public NEON tree crown dataset \cite{NEON_dataset} collected from the NEON Airborne Observation Platform surveys. This dataset provides more than $10^{8}$ crown bounding boxes for training and $10^{4}$ bounding boxes for evaluation from airborne RGB images with a spatial resolution of at least 10 cm of 37 forests across the United States. The plants over 3 m tall were defined as trees to be detected in challenging scenarios with various landforms, vegetation, climates, and ecosystem dynamics. 
To perform evaluations, three well-known object detectors with ResNet-{50} backbone including RetinaNet (i.e., pre-trained DeepForest \mbox{\cite{DeepForest}}), Faster-RCNN, and Cascade-RCNN were used. The DeepForest model was trained on over $3\times10^{7}$ algorithmically generated crown bounding boxes extracted from initial LiDAR-based tree predictions, then fine-tuned on hand-labeled bounding boxes. To effectively train the other detectors for crown detection, we initialized their backbone layers with those from the pre-trained DeepForest model and then re-trained these networks (implemented in Detectron2 \cite{Detectron2}) on NEON's training set for 80 epochs with mini-batches of size 16.
The proposed Crown-CAM is compared with three gradient-free methods, namely, the baseline Score-CAM \cite{ScoreCAM}, Augmented Score-CAM \cite{AugmentedScoreCAM}, and Eigen-CAM \cite{EigenCAM}. 
The Eigen-CAM \cite{EigenCAM} is a simple and efficient method for calculating the principle components of the learned convolutional activations, but it is not class-aware. In contrast, the Augmented Score-CAM \cite{AugmentedScoreCAM} is a computationally expensive method that uses geometric augmentations (i.e., rotation \& translation) to generate CAMs from combined activation maps of multiple input images. All experiments were conducted on an Nvidia RTX 3090 with 24GB RAM. We use images with width ($w$) and height ($h$) of 400$\times$400, while other parameters are set to $\rm{L=}$ 5, $\rm{C=}$ 256, $\rm{\bar{C}=}$ 640, $\sigma^2=$ 0.7, and $\rm{Threshold=}$ 0.4 (according to the default threshold of pre-trained DeepForest for a fair comparison and to avoid generating CAM masks with different thresholds).  \\
\begin{figure*}[!t]
\centering
\includegraphics[width=0.8\linewidth]{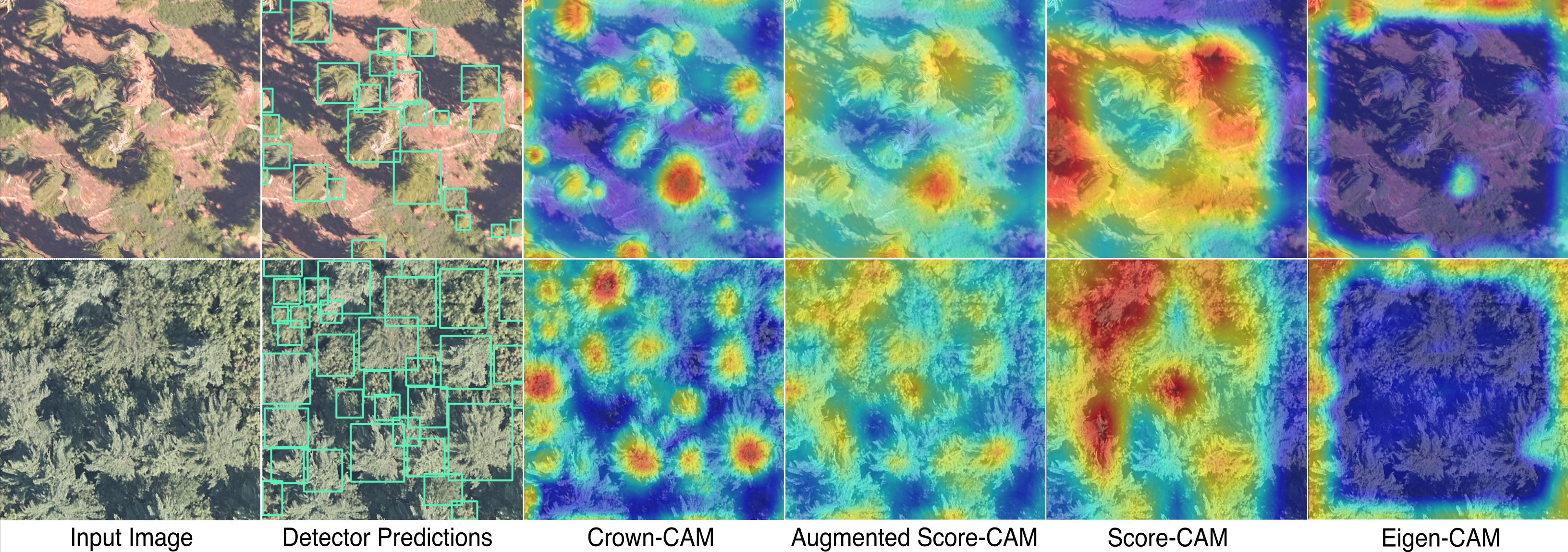}
\caption{Qualitative comparison of proposed Crown-CAM with gradient-free visual explanation methods including baseline Score-CAM \cite{ScoreCAM}, Augmented Score-CAM \cite{AugmentedScoreCAM}, and Eigen-CAM \cite{EigenCAM} for tree crown detection in aerial images. The visual explanations of the proposed Crown-CAM are highly interpretable and help to localize densely-distributed tree crowns at a finer-grain level compared to existing methods.  }
\label{fig:qual_comp}
\end{figure*}
\indent For quantitative comparison, the performance of visual explanation methods is verified on the test set of the NEON dataset using two introduced localization metrics (i.e., (\ref{eq:camiou_fg}) and (\ref{eq:camiou_bg})). Table~\ref{table:comparison} shows the average accuracy and average inaccuracy of visual explanations generated by different tree crown detectors. Accordingly, the proposed Crown-CAM outperformed other post-hoc gradient-free methods with respect to the localization effectiveness of generated CAMs, while Eigen-CAM achieved the worst results due to its inherent limitations. Note that other methods in the case of using the Cascade-RCNN model have lower \texttt{CAMIoU$^{\texttt{BG}}$} values due to their failure to generate salient regions that capture full tree crowns, thus resulting in lower intersections of generated CAMs with background regions. The proposed Crown-CAM achieves average improvements in \texttt{CAMIoU$^{\texttt{FG}}$} (and \texttt{CAMIoU$^{\texttt{BG}}$}) metric up to 8.7, 5.3, and 21.7 (and 3.3, 9.8, and 16.5) compared to the baseline Score-CAM \cite{ScoreCAM}, Augmented Score-CAM \cite{AugmentedScoreCAM}, and Eigen-CAM \cite{EigenCAM}, respectively. In addition, the CPU run-time for the Crown-CAM, Score-CAM, Augmented Score-CAM, and Eigen-CAM is 26.3, 18.4, 135.2, and 0.2 seconds per image, respectively. The results demonstrate the effectiveness of the proposed Crown-CAM in detecting tree crown areas while suppressing background regions in challenging detection scenarios involving an abundance of objects with different scales \& viewpoints and background distractions. \\
\indent Furthermore, the qualitative comparisons of visual explanations generated by different methods and the Cascade RCNN model are shown in Fig.~\ref{fig:qual_comp}. This figure illustrates two scenarios that include remotely sensed aerial images with a mixture of tree crowns \& background regions as well as densely structured forests. Based on the generated CAMs, the visual explanations of the proposed Crown-CAM are more interpretable in detecting tree crown locations compared with other existing methods that overlap irrelevant areas or result in uncertain ambiguous regions. Additionally, Fig.~\ref{fig_channel_reduction} illustrates the ablation analysis for the proposed unsupervised dimension reduction, demonstrating its negligible impact on the generated CAM while reducing computational complexity.
\section{Conclusion}\label{sec:conclusion}
In this paper, a novel interpretable visual explanation method and two localization metrics for tree crown detection in aerial images were introduced. The proposed Crown-CAM exploits unsupervised dimension reduction of activation maps to efficiently calculate local score maps emphasizing tree crown areas while suppressing background regions. Additionally, two localization metrics were derived as a function of accurate and inaccurate CAMs. These are particularly effective for analyzing errors associated with false areas of an image highlighted as salient and not carrying any contextual information about the expected tree crown detection. Empirical results of the proposed Crown-CAM demonstrated the great potential of exploring and developing interpretable XAI methods for black-box tree crown detectors that can be intuitive to human understanding of interpretability.
\section*{Acknowledgments}
\noindent The authors would like to thank Dr. Mohammad N.S. Jahromi from Aalborg University for reviewing this work and providing valuable comments and suggestions.
{\small
\bibliographystyle{IEEEtran}
\bibliography{0_references}
}

\vfill

\end{document}